\documentclass[lettersize,journal]{IEEEtran}
\usepackage{amsmath,amsfonts}

\usepackage{cite}
\usepackage{array}
\usepackage{CJKutf8} 
\usepackage{tabularx}
\usepackage{booktabs}

\usepackage{bbding}
\usepackage[caption=false,font=normalsize,labelfont=sf,textfont=sf]{subfig}
\usepackage{multirow}
\usepackage{makecell}
\usepackage{textcomp}
\usepackage{stfloats}
\usepackage{url}
\usepackage{verbatim}
\usepackage{graphicx}
\usepackage[T1]{fontenc} 
\usepackage{amssymb} 
\usepackage{algorithmic}
\usepackage{algorithm}
\usepackage[backref=False]{hyperref}
\def\BibTeX{{\rm B\kern-.05em{\sc i\kern-.025em b}\kern-.08em
    T\kern-.1667em\lower.7ex\hbox{E}\kern-.125emX}}
\usepackage{balance}
\newcommand{\Rmnum}[1]{\uppercase\expandafter{\romannumeral #1}}  
\usepackage{tikz,xcolor,hyperref}

\usepackage{amsmath}
\usepackage{subcaption}
\usepackage{graphicx}
\usepackage{amssymb}

\definecolor{lime}{HTML}{A6CE39}
\DeclareRobustCommand{\orcidicon}{
	\begin{tikzpicture}
		\draw[lime, fill=lime] (0,0)
		circle[radius=0.16]
		node[white]{{\fontfamily{qag}\selectfont \tiny \.{I}D}}; 
	\end{tikzpicture}
	\hspace{-2mm}
}
\foreach \x in {A, ..., Z}{%
	\expandafter\xdef\csname orcid\x\endcsname{\noexpand\href{https://orcid.org/\csname orcidauthor\x\endcsname}{\noexpand\orcidicon}}
}

\begin{document}
\title{Cross-Platform Chinese Offensive Comment Detection via Dual-Threshold Hard Example	Mining}

\author{Ruixing~Ren\hspace{-1.5mm}\orcidA{},  Junhui~Zhao\hspace{-1.5mm}\orcidC{},~\IEEEmembership{Senior~Member,~IEEE}, and Fangfang~Wang\hspace{-1.5mm}

\thanks{Corresponding author: Junhui Zhao.
		
Ruixing Ren, Junhui Zhao and Fangfang Wang are with the School of Electronic and Information Engineering, Beijing Jiaotong University, Beijing 100044, China. (e-mail: renruixing0604@163.com; junhuizhao@hotmail.com; 24218346@bjtu.edu.cn)

		
}
}

\maketitle

\begin{abstract}
  Cross-platform deployment of offensive comment detection for Chinese social media suffers performance degradation. The paper proposes a dual-threshold hard mining method to address this. First, the clean-Chinese-base RoBERTa is fine-tuned on COLD to establish a binary baseline for fair comparison. Second, a three-class fine-labeled test set covering Weibo, Xiaohongshu, Tieba, and Zhihu is constructed, domain distances from the source are quantified using Jaccard and Proxy-A Distance, as well as the degradation bottleneck of the baseline under domain shift is systematically revealed. Herein, a dual-threshold hard example mining strategy is proposed. High- and low-confidence error-prone samples are filtered from unlabeled corpora by prediction confidence. The model is secondarily fine-tuned under implicit contexts with merely a small set of manually labeled hard examples, realizing low-cost cross-platform domain adaptation. Experiments reveal significant performance gains of the optimized model across four platforms. 
  
\end{abstract}


\section{Introduction}
The booming of online social platforms enables free opinion expression and information sharing among netizens, yet breeds massive offensive, aggressive and hate speech \cite{SCCD,CDIAL}. Such content undermines healthy online ecosystems and triggers real-world social issues including group polarization and psychological harm \cite{Review}. Automatic detection of offensive comments on Chinese social media is an urgent research topic for network content governance and natural language processing.

Early studies mostly relied on dictionary matching or traditional machine learning, using manually constructed sensitive lexicons, TF-IDF and n-gram shallow features for discrimination. Despite high computational efficiency, such methods fail to identify implicit attacks including irony, metaphor and pun-based sarcasm in Chinese contexts \cite{COLA}. Pre-trained language models represented by BERT and RoBERTa capture deep contextual semantics via bidirectional Transformer architectures, offering new technical routes for offensive detection \cite{SWSR}. Existing benchmarks such as the COLD dataset support Chinese scenarios yet mostly confine to single platforms and static data distributions, neglecting model generalization across platforms and time domains \cite{COLD}.

Social platforms differ greatly in user demographics, linguistic styles and community jargon, causing severe performance degradation when source-domain models transfer to new platforms \cite{Ren}. Such degradation arises from domain shifts in linguistic styles, as well as strong context dependence and subjective annotation ambiguity of implicit offense. Thus, low-cost enhancement of model adaptability to heterogeneous multi-platform contexts while preserving source-domain detection performance poses urgent engineering and scientific challenges.

To address these challenges, we build a binary baseline for offensive comment detection on the public COLD dataset based on chinese-roberta-wwm-ext, validating the pre-training and fine-tuning paradigm. The main contributions are summarized as follows.
\begin{itemize}
	\item A cross-platform evaluation dataset with fine-grained three-class annotations covering four platforms is constructed, and core bottlenecks of baseline degradation under domain shifts are revealed.
	\item A lightweight domain adaptation scheme combining dual-threshold hard sample screening and implicit context fine-tuning is proposed, which substantially improves multi-platform generalization performance.
	\item Ablation experiments are performed to demonstrate that hard example mining outperforms random sampling under identical annotation budgets, verifying that sample difficulty rather than quantity dominates fine-tuning effectiveness.
\end{itemize}

The remainder of this paper is organized as follows. Related studies are reviewed in Section \ref{Sec2}. Baseline model construction and experiments are elaborated in Section \ref{Sec3}. Cross-platform evaluation and optimization strategies are discussed in Section \ref{Sec4}. Finally, conclusions are drawn and future research directions are prospected.

\section{Related Work}\label{Sec2}
Research on Chinese offensive language detection is continuously advanced with the evolution of data resources and modeling approaches. 

\subsection{Datasets and Methods for Chinese Offensive Language Detection}
Early detection methods for Chinese offensive content are mostly built on manually constructed sensitive lexicons and rule-based matching strategies. A detection and rephrasing system targeting profane language is presented in \cite{Rephrasing}, where manually designed rules are adopted to identify and rewrite frequent profane expressions. Such methods are computationally efficient yet heavily reliant on lexicon coverage, and poor generalization performance is exhibited. Traditional machine learning and deep learning approaches are subsequently introduced by researchers. A hybrid CNN-Bi-GRU model integrating character-level and word-level features is proposed by Zhang et al. \cite{Zhang} to mitigate the impact of word segmentation errors on semantic comprehension of short texts.

The development of this field is significantly promoted by the emergence of pre-trained language models. The Chinese offensive language detection benchmark COLD is constructed by Deng et al. \cite{COLD}. Data of this dataset are collected from Zhihu and Weibo, covering three themes including race, gender and region, and a detector named COLDetector is obtained via fine-tuning based on bert-base-chinese. ToxiCN is further proposed by Lu et al. \cite{Facilitating}. Samples of the dataset are collected from Zhihu and Baidu Tieba. A hierarchical annotation framework that distinguishes general offensive content from hate speech is established, and coverage of implicit and reported expressions is enhanced. Modeling methods are continuously improved in follow-up works based on resources such as COLD. A three-level feature fusion model integrating character, word and sentence features is proposed by Li et al. \cite{Offensive} to boost the recognition of offensive texts with complex semantics. RoBERTa is combined with pointer networks by Hou et al. \cite{Hou}, and attention is focused on offensive-related critical fragments. A two-channel approach integrating BERT and topic models is proposed by Cao et al. \cite{Cao} to simultaneously capture semantic and topic features. The effectiveness of the paradigm consisting of large-scale pre-training and downstream fine-tuning for Chinese offensive detection is verified by these studies. This paradigm is also followed by the baseline model of this paper.

\subsection{Recognition of Implicit Offense and Context-Dependent Attacks}

Attacks with explicit abusive words are contrasted with implicit offenses disguised by irony, metaphor, homophones. No violating words are contained in their literal expressions, yet strong offensive intentions are implied, and major difficulties in detection are caused accordingly. Corpus construction and linguistic pattern modeling are devoted to by numerous existing studies in the field of irony detection. The construction of traditional Chinese irony corpora and the analysis of ironic structures are completed in early research \cite{Irony}. Neural network methods integrating linguistic features are proposed subsequently \cite{Lu}. Retrospective reading models emphasizing contextual information and sarcasm recognition approaches incorporating news background are further developed \cite{Novel}\cite{Gong}. It is demonstrated by all these works that the recognition of implicit expressions is highly dependent on context modeling. A multi-domain Chinese implicit hate dataset MCIHD is constructed by recent research \cite{Domain}, and domain-enhanced prompt learning is put forward for implicit hate and euphemistic expressions. TE-Dataset targeting harmful euphemisms together with its contrastive learning framework is designed to specially cope with evasion methods such as metaphor, homophones and character deformation \cite{Euphemism}. It is demonstrated by all these studies that the recognition of implicit offenses needs to go beyond keyword matching and delve into contextual and semantic correlations. Consistency with the motivation of this paper to strengthen implicit context modeling via hard example mining is therefore achieved.

\subsection{Cross-Platform Generalization and Model Robustness}
In practical deployment, target platforms with data distributions different from training data are often faced by models, and their generalization ability and robustness are thus concerned by researchers. Generalization performance is improved by one category of works through data augmentation. A million-scale unsupervised dataset AugCOLD is constructed \cite{Enhancing}, and a multi-teacher distillation framework is proposed. Coverage of hard samples including implicit offenses and retaliatory remarks is significantly expanded, and the robustness of the detector is enhanced accordingly. Anti-evasion robustness evaluation is focused on by another line of research. Perturbed samples are constructed via homophone replacement and emoji substitution in ToxiCloakCN \cite{ToxiCloakCN}. Structural perturbations such as character splitting and radical replacement are introduced in CangjieToxi \cite{CangjieToxi}. Significant performance degradation on perturbed samples is found in both datasets for existing models, and implicit offenses related to cultural backgrounds can hardly be understood. The vulnerability of models trained on source domains under distribution shift is revealed by these studies. However, most improvements are realized by relying on large-scale data augmentation or optimization for specific perturbation types. Systematic quantification of domain gaps under real multi-platform scenarios and target-platform adaptation with minimal labeling cost are rarely achieved in existing works.

\subsection{Positioning of This Work}
Remarkable progress in data resources, implicit offense modeling and robustness evaluation has been achieved by existing studies. Nevertheless, three deficiencies still exist: First, evaluations are conducted on a single platform or static distribution in most works, and systematic quantification of real-world cross-platform domain shift is absent. Second, although implicit offenses have attracted widespread attention, detailed diagnosis of their failure mechanisms during cross-platform transfer is lacking. Third, most schemes for boosting robustness are realized by large-scale data augmentation, and high labeling costs are incurred accordingly.

To address these limitations, a cross-platform evaluation dataset covering four platforms with three classification categories is constructed in this paper. Domain distances between each platform and the source domain COLD are quantified. Meanwhile, a lightweight domain adaptation method based on dual-threshold hard sample mining is proposed. The cross-platform recognition performance of implicit offenses is selectively improved with an extremely small labeling budget.

\section{Construction of RoBERTa-Based Baseline Model for Chinese Offensive Comment Detection} \label{Sec3}
To obtain unbiased and fairly comparable fine-tuning results, the clean Chinese pre-trained backbone chinese-roberta-wwm-ext is adopted. Fine-tuning is conducted from scratch on the COLD training set with binary cross-entropy loss, and the baseline detector is obtained consequently.

\begin{table}[t]
	\centering
	\caption{Sample Statistics of COLD Dataset}
	\renewcommand{\arraystretch}{1.3}
	\label{tab1}
	\begin{tabular}{lccc}
		\hline
		Dataset Subset & Offensive Samples & Normal Samples & Total \\
		\hline
		Train & 12723 & 13003 & 25726 \\
		Validation & 3211 & 3220 & 6431 \\
		Test & 2107 & 3216 & 5323 \\
		\hline
		Total & 18041 & 19439 & 37480 \\
		\hline
	\end{tabular}
\end{table}
\subsection{Selection and Processing of Experimental Datasets}
The COLD \cite{COLD} is selected as the core experimental data. It is specially constructed for harmful, offensive and malicious speech in Chinese online. Mainstream social platforms including Zhihu and Weibo are covered, and controversial topics such as race, gender and region are included. Better generalization ability of the model is facilitated accordingly. The sample distribution of each subset of the COLD dataset is shown in Table \ref{tab1}.

\subsection{Text Preprocessing and Vectorization}
Given the raw text sequence $S$, regular expression cleaning is implemented first. Only Chinese characters within the Unicode range [$\setminus$u4e00,$\setminus$u9fa5] are retained, while irrelevant characters including punctuations, symbols, digits and foreign languages are eliminated. The pure Chinese character sequence after cleaning is defined as:
\begin{equation}
	S_{\text{cleaned}} = f_{\text{regex}}(S) = \bigl\{c \,\big|\, c \in S,\ c \in [\backslash\text{u4e00},\backslash\text{u9fa5}]\bigr\}
\end{equation}
Subsequently, word segmentation is performed on the cleaned string via the segmentation operator Cut($\cdot$) based on prefix dictionaries and Hidden Markov Models, and the discrete word segmentation set $W$ is obtained:
\begin{equation}
	W = \{w_1,w_2,\dots,w_n\} = \mathrm{Cut}(S_{\mathrm{cleaned}})
\end{equation}
Stop word removal is further carried out, where $V_{\text{stop}}$ denotes the predefined set of Chinese stop words. Double filtering is applied to the word segmentation set $W$: stop words and single-character words with length $\le1$ are discarded. Words satisfying $w \notin V_{\mathrm{stop}} \land \mathrm{Length}(w) > 1$ are retained to form the final valid feature sequence:
\begin{equation}
	W_{\mathrm{final}} = \bigl\{w \in W \,\big|\, w \notin V_{\mathrm{stop}} \land \mathrm{Length}(w) > 1\bigr\}
\end{equation}
The indicator function $\mathbf{I}(\cdot)$ is adopted to count the occurrence frequency of each word $w$ in $W_{\text{final}}$, and the indicator function satisfies:
\begin{equation}
	\textbf{I}(A)=
	\begin{cases}
		1, & \text{if condition } A \text{ holds} \\
		0, & \text{if condition } A \text{ does not hold}
	\end{cases}
\end{equation}
The word frequency calculation formula of vocabulary $w$ is given as:
\begin{equation}
	\mathrm{TF}(w) = \sum_{i=1}^{|W_{\mathrm{final}}|} \mathbf{I}(W_{\mathrm{final},i}=w)
\end{equation}
An evolution example of one sample through the above procedures is illustrated in Table \ref{tab2}.
\begin{table*}[t]
	\centering
	\caption{Example of Social Comment Evolution in Each Preprocessing Stage}
	\renewcommand{\arraystretch}{1.3}
	\begin{tabular}{|l|l|l|}
		\hline
		Processing Stage & Data Status & Text Features and Output Results \\
		\hline
		Raw Input Text $S$ & Raw long text with symbols & \begin{CJK}{UTF8}{gbsn}都怪我我应该送花圈。这样以后也能用得上\end{CJK} \\
		\hline
		Regex Cleaning $S_{\mathrm{cleaned}}$ & Pure Chinese sequence without punctuation & \begin{CJK}{UTF8}{gbsn}都怪我我应该送花圈这样以后也能用得上\end{CJK} \\
		\hline
		Word Segmentation $W=\mathrm{Cut}(S_{\mathrm{cleaned}})$ & Full segmented word set & \{\begin{CJK}{UTF8}{gbsn}都, 怪, 我, 我, 应该, 送, 花圈, 这样, 以后, 也, 能, 用得上\end{CJK}\} \\
		\hline
		Stopword \& Length Filter $W_{\mathrm{final}}$ & Filtered core feature words & \{\begin{CJK}{UTF8}{gbsn}应该, 送, 花圈, 这样, 以后, 用得上\end{CJK}\} \\
		\hline
		Term Frequency Calculation $\mathrm{TF}(w)$ & Word-TF mapping & \begin{CJK}{UTF8}{gbsn}应该: 1, 送: 1, 花圈: 1, 这样: 1, 以后: 1, 用得上: 1 \end{CJK}\\
		\hline
	\end{tabular}
	\label{tab2}
\end{table*}

To feed the preprocessed text into deep neural networks, subword segmentation and numerical mapping are further performed to generate tensors with fixed dimensions. It is worth noting that character-level subword segmentation is executed by the built-in Tokenizer of RoBERTa. Special tokens $[\text{CLS}]$ and $[\text{SEP}]$ are appended to the start and end of the sequence respectively, and an ordered discrete token sequence is constructed accordingly, as follows.
\begin{equation}
	T = \left[t_{\mathrm{CLS}}, t_1, t_2, \dots, t_n, t_{\mathrm{SEP}}\right],\quad L = n+2
\end{equation}
where $n$ denotes the number of valid tokens in the original text, and $L$ stands for the total sequence length after inserting special tokens. The hidden layer vector corresponding to the first position $t_{\text{CLS}}$ is adopted to aggregate the global semantic information of the entire sentence. $t_{\text{SEP}}$ is treated as the terminal boundary token to mark the end of a single text.

Subsequently, with the aid of the vocabulary built into the model, each Token in the sequence is mapped to a unique integer index in the vocabulary via the lookup operator $f_{\text{lookup}}(\cdot)$, and a one-dimensional index vector $\textbf{X}_{\text{ids}}\in \mathbb{Z}^L$ is obtained:
\begin{equation}
	\mathbf{X}_{\mathrm{ids}} = [f_{\mathrm{lookup}}(t_\text{CLS}),f_{\mathrm{lookup}}(t_i)\mid_{i=1,...,n},f_{\mathrm{lookup}}(t_\text{SEP})] 
\end{equation}
Finally, the batch dimension is expanded by the tensor reshaping operator $f_{\text{tensor}}$. The index vectors of individual samples are stacked to obtain the batch input tensor $\textbf{X}$ that can be fed into the neural network, as follows.
\begin{equation}
	\textbf{X} = f_{\mathrm{tensor}}(\textbf{X}_{\mathrm{ids}}) \in \mathbb{Z}^{b\times L}
\end{equation}
where $b$ is defined as the batch size for model training.

\begin{figure}[t]
	\centerline{\includegraphics[width=3.6in,keepaspectratio]{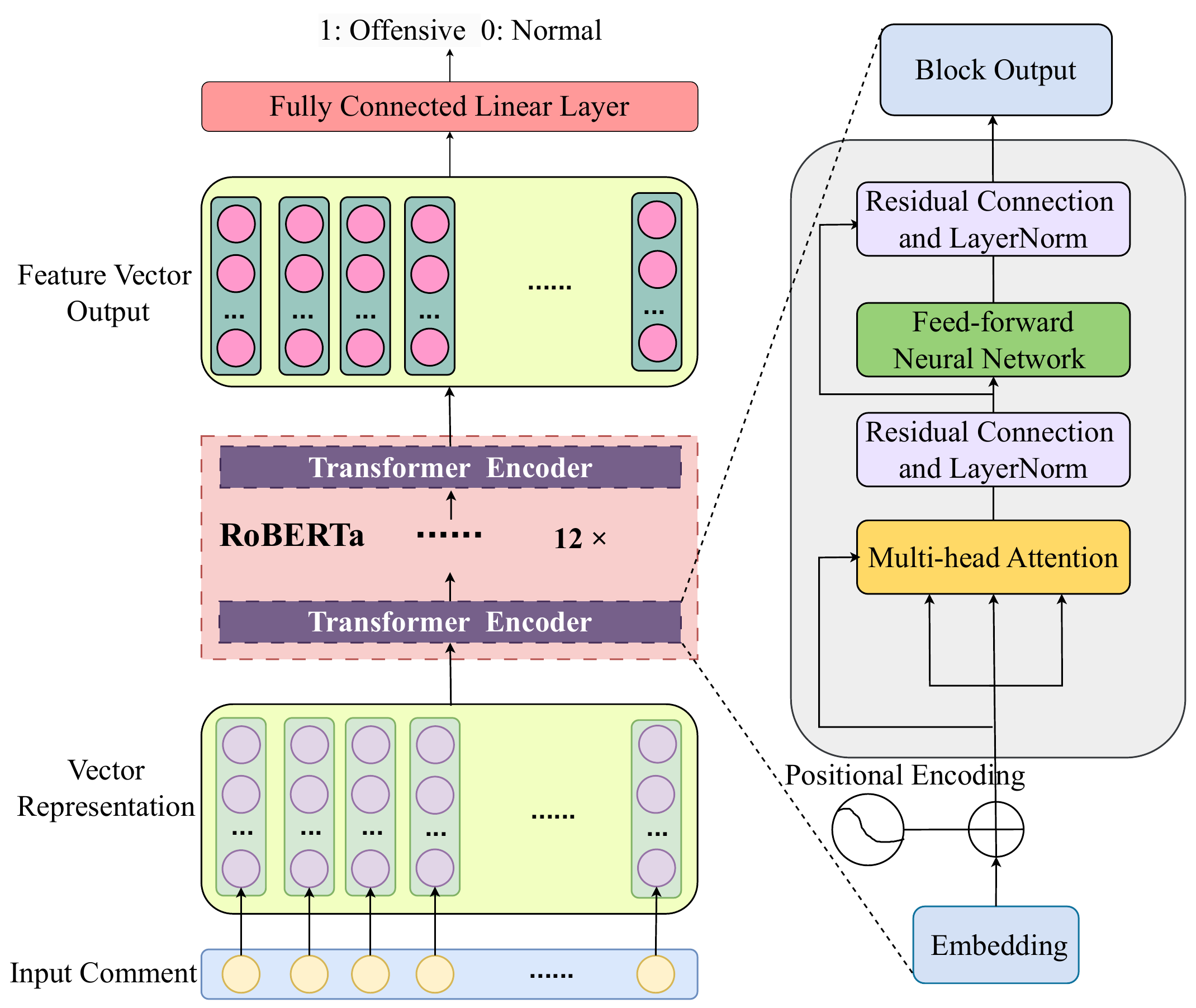}}
	\caption{Network Architecture of RoBERTa for Text Classification}
	\label{fig1}
\end{figure}

\subsection{Design of Classification Network}
The batch input tensor $\textbf{X}\in\mathbb{Z}^{b\times L}$ is obtained through the preprocessing procedures described above. Global semantic extraction is implemented by the bidirectional Transformer encoder of RoBERTa in this section, and a binary classification head is constructed. The overall network architecture is illustrated in Fig. \ref{fig1}.

\textbf{Global Semantic Feature Vector Extraction.}
The pre-trained model chinese-roberta-wwm-ext adopted in this paper follows the RoBERTa architecture. Multi-layer bidirectional Transformer encoders are taken as the backbone for feature extraction, whose core component is multi-head self-attention mechanism. The model is abbreviated as RoBERTa in the following text. Linear projections are separately performed on input features by this mechanism to generate query matrix $\mathbf{Q}$, key matrix $\mathbf{K}$ and value matrix $\mathbf{V}$. The calculation formula of scaled dot-product attention is given as follows:
\begin{equation}
	\mathrm{Attention}(\mathbf{Q},\mathbf{K},\mathbf{V}) = \mathrm{softmax}\left(\frac{\mathbf{Q}\mathbf{K}^\mathrm{T}}{\sqrt{d_k}}\right)\mathbf{V}
\end{equation}
where $d_k$ denotes the feature dimension of key vectors, and $\sqrt{d_k}$ is adopted as the scaling factor to constrain the value of inner product and avoid gradient vanishing. $h$ groups of mutually independent linear projections and attention calculations are conducted on features by multi-head attention. The outputs of each head are concatenated and fused via weight matrix $\mathbf{W}^O$:
\begin{equation}
	\mathrm{MultiHead}(\mathbf{Q},\mathbf{K},\mathbf{V})=\mathrm{Concat}(\mathrm{head_1},...,\mathrm{head_h})\mathbf{W}^O
\end{equation}
\begin{equation}
	\mathrm{head_i}=\mathrm{Attention}(\mathbf{Q}\mathbf{W}_i^\mathrm{Q},\mathbf{K}\mathbf{W}_i^\mathrm{K},\mathbf{V}\mathbf{W}_i^\mathrm{V})
\end{equation}

As illustrated in Fig. \ref{fig1}, 12 layers of the above Transformer encoder blocks are stacked in RoBERTa. Benefiting from the bidirectional encoding property, contextual semantic information on both left and right sides can be fused simultaneously by any Token within the sequence. After tensor $\mathbf{X}$ is fed into the RoBERTa encoder, a feature matrix with global context awareness is output:
\begin{equation}
	\begin{aligned}
		\mathbf{H} &= \mathrm{RoBERTa}(\mathbf{X}) =\\ &\bigl[\mathbf{h}_{\mathrm{CLS}}, \mathbf{h}_1, \mathbf{h}_2, \dots, \mathbf{h}_n, \mathbf{h}_{\mathrm{SEP}}\bigr] \in \mathbb{R}^{b\times L\times d_{\mathrm{hidden}}}
	\end{aligned}
\end{equation}
where $d_\mathrm{hidden}$ denotes the hidden feature dimension of the model. The feature vector $\mathbf{h}_i$ at arbitrary position $i$ is jointly mapped from the semantics of all Tokens in the entire sentence, which is formulated as $\mathbf{h}_i=g(T)$. The token $[\text{CLS}]$ is endowed with the ability to aggregate the global semantics of the whole text during the pre-training stage of RoBERTa. For the binary classification task of single offensive text in this paper, only the feature vector at the first position $\mathbf{h}_{\mathrm{CLS}}\in \mathbb{R}^{d_{\text{hidden}}}$ is extracted as the global semantic representation of the entire comment text. $\mathbf{h}_1, \mathbf{h}_2, \dots, \mathbf{h}_n$ in the middle of the sequence are local semantic features of individual words, and the terminal vector $\mathbf{h}_{\mathrm{SEP}}$ is only used to mark sentence boundaries. Since this paper does not target word-level offensive localization tasks, neither of them is involved in the subsequent calculation for sentence-level classification.

\textbf{Design of Classification Head.}
The classification module is composed of a single fully-connected linear layer cascaded with the Softmax normalization function. The mapping from $d_\mathrm{hidden}$-dimensional global features to the binary classification probability space is realized. The extracted global vector $\mathbf{h}_{\text{CLS}}$ is fed into the fully-connected layer. The raw scores of each category are calculated via learnable weight matrix $\mathbf{W}$ and bias term $\mathbf{b}$:
\begin{equation}
	\mathbf{z} = \mathbf{W} \cdot \mathbf{h}_{\mathrm{CLS}} + \mathbf{b},\quad \mathbf{W} \in \mathbb{R}^{2\times d_{\mathrm{hidden}}},\ \mathbf{b} \in \mathbb{R}^{2}
\end{equation}
The raw scores are then normalized into category probabilities within the interval $[0,1]$ via the Softmax activation function:
\begin{equation}
	p_i = \frac{e^{z_i}}{\sum_{j=0}^{1}e^{z_j}},\quad z_i\in \mathbf{z},i\in\{0,1\}
\end{equation}
where $i=0$ denotes normal non-offensive texts, and $i=1$ denotes offensive harmful texts. The label corresponding to the maximum probability is finally selected via the $\mathrm{argmax}$ function as the prediction result of the model:
\begin{equation}
	\hat{y} = \mathop{\arg\max}_{i\in\{0,1\}} p_i
\end{equation}

\subsection{Model Fine-tuning and Training Configuration}
Windows 11 is adopted as the operating system in all experiments of this paper, and NVIDIA GeForce RTX 4060 Ti with 16 GB video memory is used as the GPU. Python 3.8.19 serves as the programming environment, PyTorch 2.3.0 is selected as the deep learning framework, and CUDA 12.1 is configured for GPU acceleration. To precisely control the convergence of the large-scale pre-trained model, intensive fine-tuning is carried out with training steps as the core metric in this experiment. The relevant hyperparameter settings are presented in Table \ref{tab3}. The parameter count of each component and the total number of floating-point operations of the entire model are listed in Table \ref{tab4}.

\begin{table}[htbp]
	\centering
	\caption{Model hyperparameters}
		\renewcommand{\arraystretch}{1.3}
	\begin{tabular}{cc}
		\hline
		Hyperparameter & Value \\
		\hline
		Hidden dimension & 768 \\
		Learning rate & $2\times10^{-5}$ \\
		Optimizer & AdamW \\
		Maximum sequence length & 128 \\
		Weight decay & 0.005 \\
		\hline
	\end{tabular}
	\label{tab3}
\end{table}

\begin{table}[htbp]
	\centering
	\caption{Model parameter counts and computational costs}
		\renewcommand{\arraystretch}{1.3}
	\begin{tabular}{ccc}
		\hline
		Module & Parameters & FLOPs \\
		\hline
		RoBERTa encoder & 102.2M & 5.44B \\
		Linear classifier & 0.001M & 0.0001B \\
		\hline
	\end{tabular}
	\label{tab4}
\end{table}

Binary cross-entropy loss is used for fine-tuning supervision:
\begin{equation}
	\mathcal{L} = -\frac{1}{N}\sum_{i=1}^{N}\Big[y_i\log p_i^{(1)} + (1-y_i)\log\big(1-p_i^{(1)}\big)\Big]
\end{equation}
where $N$ denotes batch size, $y_i\in \{0,1\}$ denotes ground-truth label, and $p_i^{(1)}$ denotes the predicted probability of harmful text. AdamW optimizer is adopted with the following update rule:
\begin{equation}
	\Theta^{(t+1)} = \Theta^{(t)} - \eta\nabla_{\Theta}\mathcal{L}
\end{equation}
where $\eta=2\times10^{-5}$, with weight decay of 0.01 to mitigate overfitting.

\begin{figure}[t]
	\centerline{\includegraphics[width=3.5in,keepaspectratio]{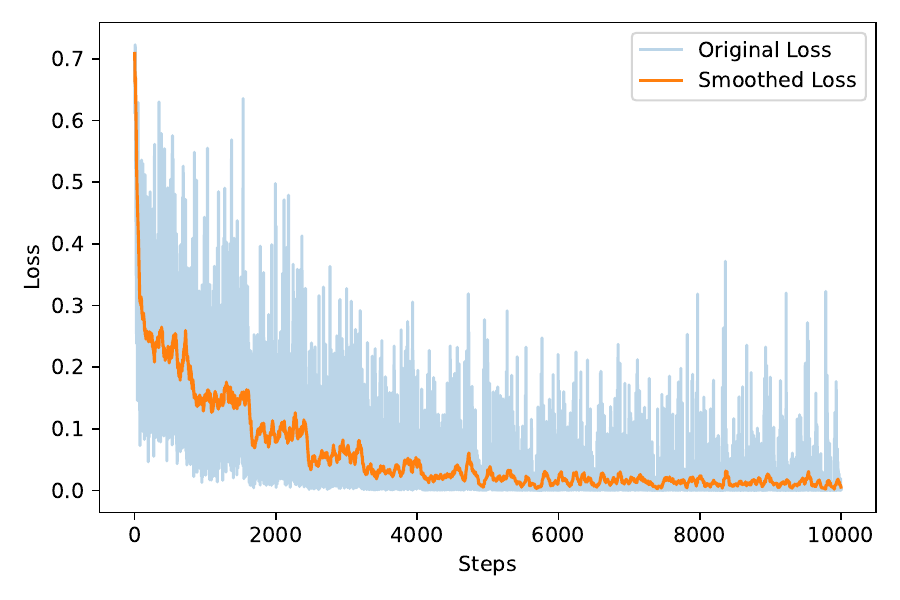}}
	\caption{Training loss convergence curve}
	\label{fig2}
\end{figure}

\subsection{Analysis of baseline experimental results}
Fig. \ref{fig2} shows training loss convergence. Loss declines rapidly: it drops from $\sim$0.7 to below 0.1 within the first 2000 steps and stabilizes near zero after 4000 steps, indicating full convergence on the COLD.

\begin{figure}[t]
	\centerline{\includegraphics[width=3.6in,keepaspectratio]{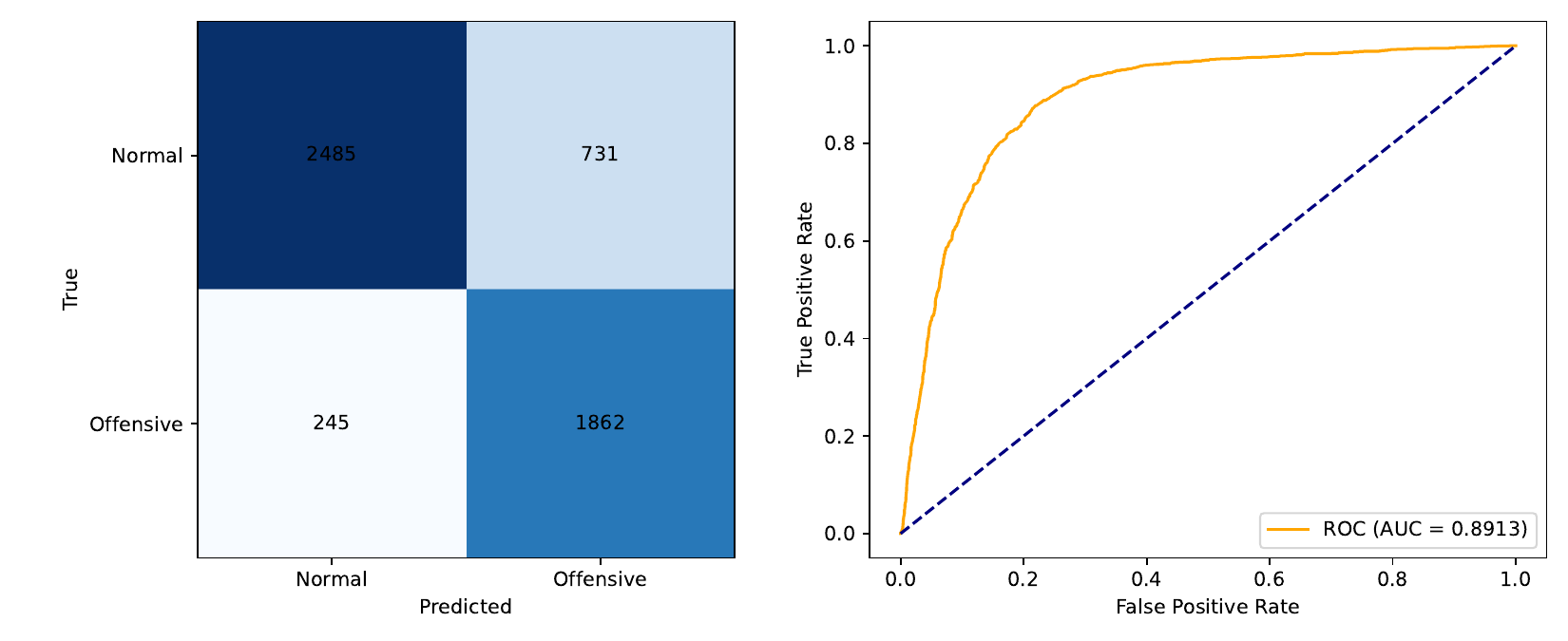}}
	\caption{Confusion matrix and ROC curve on test set}
	\label{fig4}
\end{figure}

Confusion matrix and ROC on the test set are presented in Fig. \ref{fig4}. The matrix reveals 245 false-negative offensive samples and 731 false-positive normal samples. The model achieves higher recall for offensive content to minimize missed harmful texts. The ROC AUC reaches 0.891, substantially exceeding the random baseline, demonstrating strong discriminative power between positive and negative samples.

\begin{table}[htbp]
	\centering
	\caption{Performance comparison of different models}
			\renewcommand{\arraystretch}{1.3}
	\begin{tabular}{lccccc}
		\hline
		Model & Accuracy & Precision & Recall & F1 & AUC \\
		\hline
		LR & 0.620 & 0.559 & 0.190 & 0.284 & 0.567 \\
		SVM & 0.625 & 0.579 & 0.190 & 0.286 & 0.570 \\
		RF & 0.621 & 0.579 & 0.155 & 0.245 & 0.564 \\
		RoBERTa & 0.842 & 0.782 & 0.884 & 0.792 & 0.891 \\
		\hline
	\end{tabular}
	\label{tab5}
\end{table}

To verify RoBERTa’s superiority in deep semantic modeling, logistic regression, SVM and RF are adopted as baselines on the identical test set; evaluation metrics are listed in Table \ref{tab5}. The three traditional models achieve $\sim$0.62 accuracy, F1 scores of 0.25-0.29 and offensive-sample recall near 0.17. Reliant on sparse TF-IDF and N-gram bag-of-word features, they only capture superficial lexical signals and fail to detect context-dependent offensive language such as sarcasm and metaphor. Imbalanced data further biases these conservative classifiers toward predicting most samples as non-offensive. By contrast, RoBERTa encodes global semantics via bidirectional Transformers, yielding an F1 improvement of 0.508 over logistic regression and capturing subtle hostile intentions accurately. This validates the efficiency of large Chinese corpus pre-training followed by downstream fine-tuning for offensive comment detection.

\section{Evaluation and optimization of cross-platform generalization} \label{Sec4}
This section evaluates the transferability of the baseline detector across platforms, and optimizes few-shot domain adaptation via dual-threshold hard example mining.

\subsection{Cross-platform test set construction and annotation quality}
The cross-platform evaluation dataset consists of two subsets. (1) Crawled comment data from trending topics on Weibo and Xiaohongshu; (2) Aggregated adversarial, long and implicit toxic texts from Tieba and Zhihu within the ToxiCN dataset \cite{Facilitating}. All data underwent three-stage preprocessing (structural noise removal, semantic purification, global deduplication) to yield clean corpora.

\begin{table}[htbp]
	\centering
	\caption{Sample distribution of the cross-platform test set}
	\renewcommand{\arraystretch}{1.3}
	\begin{tabular}{lccc}
		\hline
		Platform & 0 & 1 & 2 \\
		\hline
		Weibo & 337 & 80 & 83 \\
		Xiaohongshu & 332 & 24 & 144 \\
		Tieba & 251 & 123 & 126 \\
		Zhihu & 270 & 87 & 143 \\
		Total & 1190 & 314 & 496 \\
		\hline
	\end{tabular}
	\label{tab6}
\end{table}

To accurately distinguish different offensive patterns and refine the analysis of generalization errors during annotation, a refined three-class labeling criterion is established: 0 denotes normal comments; 1 denotes explicit offensiveness, including overt insults, vulgarities and discriminatory content with exposed aggressive semantics; 2 denotes implicit offensiveness, which adopts sarcasm, metaphor and homophonic innuendoes to conceal hostile intentions without explicit toxic words. Stratified random sampling of 500 samples per platform is conducted from full corpora. A double-blind cross-labelling scheme with third-party arbitration is used to build an independent cross-platform test set. This test set is excluded from all training and parameter tuning procedures. Sample distributions across platforms are shown in Table \ref{tab6}.

\textbf{Inter-annotator agreement.} To guarantee reliable evaluation, we report the inter-annotator agreement. The overall Cohen’s/Fleiss’ Kappa reaches 0.55 with a 77.5\% agreement rate, with obvious discrepancies across platforms: Zhihu achieves the highest agreement (92.5\%), while Tieba yields the lowest (62.5\%). These results reveal inherent subjectivity in identifying implicit offensive content, alongside larger mismatches between texts from platforms such as Tieba and the original labeling standards. This motivates separate evaluation of implicit offensive samples and dedicated discussion of labeling inconsistencies.

\textbf{Error auditing protocol.} To quantify labeling error rates on the test set, stratified random sampling by platform and label was performed on samples with mismatches between baseline predictions and manual annotations. Each sample was manually categorized as model misprediction, incorrect label, or ambiguous borderline case. A total of 192 discrepant samples are audited manually: 114 (59.4\%) are model mispredictions, 43 (22.4\%) are ambiguous borderline cases, and only 35 (18.2\%) carry incorrect labels. Nearly six-tenths of inconsistencies stem from model limitations rather than annotation flaws, with genuine labeling errors accounting for merely around one-fifth. This result validates the high annotation quality of the cross-platform test set. As a preliminary sanity check, an additional set of 106 stratified samples (with oversampling of implicit offensive instances) is reviewed by a single annotator. Around 97\% matched the original labels, while clear labeling errors accounted for merely 3\%, with disagreements concentrated on the borderline between implicit offensiveness and normal comments. 

\begin{table}[t]
	\centering
	\renewcommand{\arraystretch}{1.3}
	\caption{Quantification of domain discrepancies across platforms}
	\begin{tabular}{c c c c c c}
		\hline
		Proxy-A$\setminus$Jaccard	& Weibo & Xiaohongshu & Tieba & Zhihu & COLD \\
		\hline
		Weibo & --- & 0.219 & 0.216 & 0.215 & 0.209 \\
		Xiaohongshu & 0.453 & --- & 0.168 & 0.172 & 0.173 \\
		Tieba & 0.91 & 1.06 & --- & 0.357 & 0.344 \\
		Zhihu & 1.24 & 1.33 & 0.85 & --- & 0.385 \\
		COLD & 1.67 & 1.69 & 1.32 & 0.96 & --- \\
		\hline
	\end{tabular}
	\label{tab7}
\end{table}

\textbf{Quantification of platform discrepancies.} To objectively characterize cross-platform discrepancies instead of relying on subjective intuition, we adopt vocabulary Jaccard overlap and Proxy-A-distance as evaluation metrics. The smaller the former, the greater the difference; the larger the latter, the greater the difference. Quantitative results are summarized in Table \ref{tab7}. Xiaohongshu exhibits the largest domain gaps against all other platforms, while Tieba and Zhihu share the closest distribution, presumably due to their shared source within the ToxiCN dataset. We further compute each platform’s relative distance to the COLD dataset to directly measure transfer difficulty. Zhihu lies closest to COLD, followed by Tieba, whereas Xiaohongshu and Weibo show the largest domain gaps from COLD. This indicates that Xiaohongshu, a platform entirely absent from the COLD dataset, has the largest domain gap with the source domain. Although Weibo appears in COLD, its actual linguistic distribution deviates substantially, likely driven by differences in collection time, topic coverage and data preprocessing pipelines.

\begin{table}[t]
	\centering
	\renewcommand{\arraystretch}{1.3}
	\caption{Generalization performance statistics of the baseline model across social platforms}
	\begin{tabular}{lcccc}
		\hline
		Platform & Accuracy & Precision & Recall & F1-Score \\
		\hline
		Weibo & 0.708 & 0.591 & 0.337 & 0.430 \\
		Xiaohongshu   & 0.652 & 0.444 & 0.143 & 0.216 \\
		Tieba & 0.592 & 0.613 & 0.490 & 0.545 \\
		Zhihu     & 0.618 & 0.595 & 0.530 & 0.561 \\
		\hline
	\end{tabular}
	\label{tab8}
\end{table}
\begin{figure}[t]
	\centerline{\includegraphics[width=3.6in,keepaspectratio]{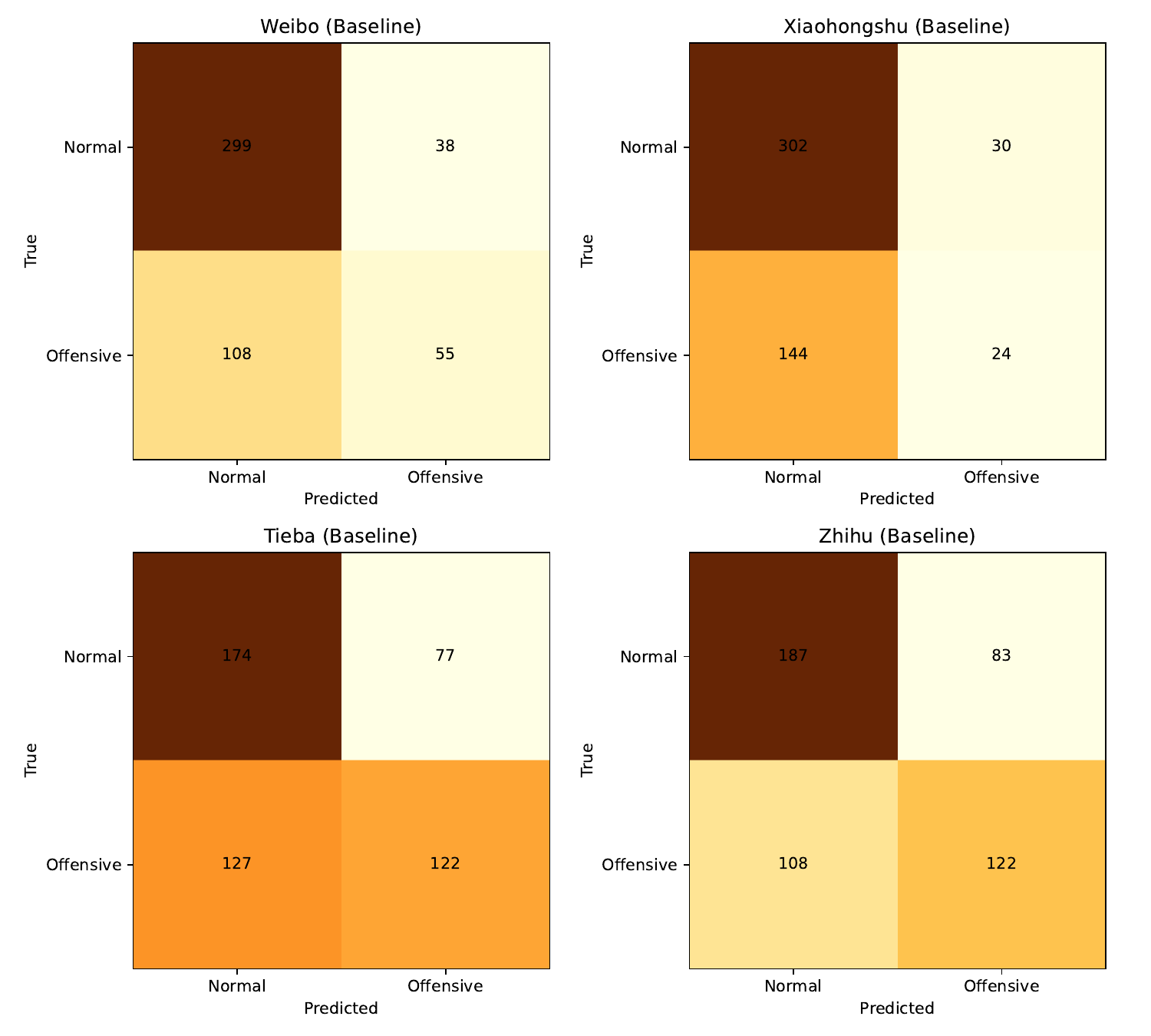}}
	\caption{Cross-domain confusion matrices of the baseline detector for each platform}
	\label{fig5}
\end{figure}
\subsection{Cross-platform performance of the baseline detector}
The baseline detector is directly inferred on the four test subsets. Metrics for each platform are listed in Table \ref{tab8}, and confusion matrices are illustrated in Fig. \ref{fig5}. The results reveal obvious overall performance degradation of the detector on cross-platform data compared with the source domain, with the most severe drop observed on Xiaohongshu.

Combined with the confusion matrices in Fig. \ref{fig5}, cross-domain transfer failure mainly stems from two factors. First, the model exhibits weak detection capacity for implicit offensive content. A large number of implicit offensive samples conveyed via irony and metaphor are falsely classified as normal, which constitutes the primary cause of low recall. This issue is most prominent on Xiaohongshu, where 144 out of 168 offensive samples are missed. The false negative counts for Zhihu and Tieba reach 108 and 127 respectively. Second, cross-domain transfer fails for platform-exclusive jargon. Some offensive texts contain no sensitive words from conventional swear word lexicons and deliver implicit insults through platform-specific slang, while the source domain lacks analogous training instances. This observation aligns with the analysis in Section \ref{Sec4} A. As a completely unseen domain with the largest distance from COLD, Xiaohongshu suffers the most severe performance degradation, consistent with the expectation that larger domain gaps lead to poorer transferability.

\subsection{Dual-Threshold Hard Example Mining and Implicit Context Model Calibration}
This section designs a dual-threshold dual-channel hard sample screening strategy to mine cross-domain hard examples, and perform secondary fine-tuning calibration on implicit context based on the screened hard samples.

\textbf{(1) Dual-Threshold Hard Example Mining}

To accurately capture samples where the baseline model fails during cross-platform inference, we establish a dual-threshold hard sample screening rule based on prediction confidence. Hard examples prone to model misclassification are mined from massive unlabeled candidate corpora and adopted as fine-tuning data for subsequent model calibration. 

Since the well-trained baseline model follows a binary classification setup, label space alignment is required as a preliminary step. The original human annotations are three-class ground-truth labels $y_{\mathrm{human}}\in \{0,1,2\}$. We integrate them into binary training labels $y_{\mathrm{binary}}$ via a mapping operator, unifying explicit and implicit offensive instances into a single offensive class. The conversion rule is formulated as follows:
\begin{equation}
	y_{\mathrm{binary}}=
	\begin{cases}
		1 & y_{\mathrm{human}}\ge 1 \\
		0 & y_{\mathrm{human}} = 0
	\end{cases}
\end{equation}
The baseline model outputs class-wise confidence scores normalized by Softmax. The maximum probability is taken as the per-sample decision confidence $c$, whose calculation is defined in the following formula:
\begin{equation}
	c = \max_{k\in\{0,1\}} \left( \frac{e^{z_k}}{e^{z_0}+e^{z_1}} \right)
\end{equation}
where $z_0$ and $z_1$ denote the raw logits of the non-offensive and offensive classes, respectively. Accordingly, the confidence score $c$ satisfies $c\in[0.5,1.0]$.

On the condition that the model predicted label $\hat{y}$ on the self-built dataset disagrees with the binary ground-truth label $y_{\mathrm{binary}}$, two hard sample pools are divided according to confidence intervals. We set the high-confidence threshold $\alpha$ and the low-confidence threshold $\beta$:
\begin{enumerate}
	\item \textbf{High-Confidence Error Pool:} $\hat{y} \neq y_{\mathrm{binary}} \land c \ge \alpha$. Most samples in this pool are implicit offensive texts concealed by irony and metaphor with amiable appearances, which are falsely predicted as normal with high confidence by the model.
	\item \textbf{Low-Confidence Error Pool:} $\hat{y} \neq y_{\mathrm{binary}} \land 0.50 \le c \le \beta$. Most samples in this pool contain platform-specific slang and emerging internet buzzwords. The model presents ambiguous decision boundaries for such texts, with prediction confidence fluctuating around 0.5.
\end{enumerate}

\begin{figure}[t]
	\centerline{\includegraphics[width=3.0in,keepaspectratio]{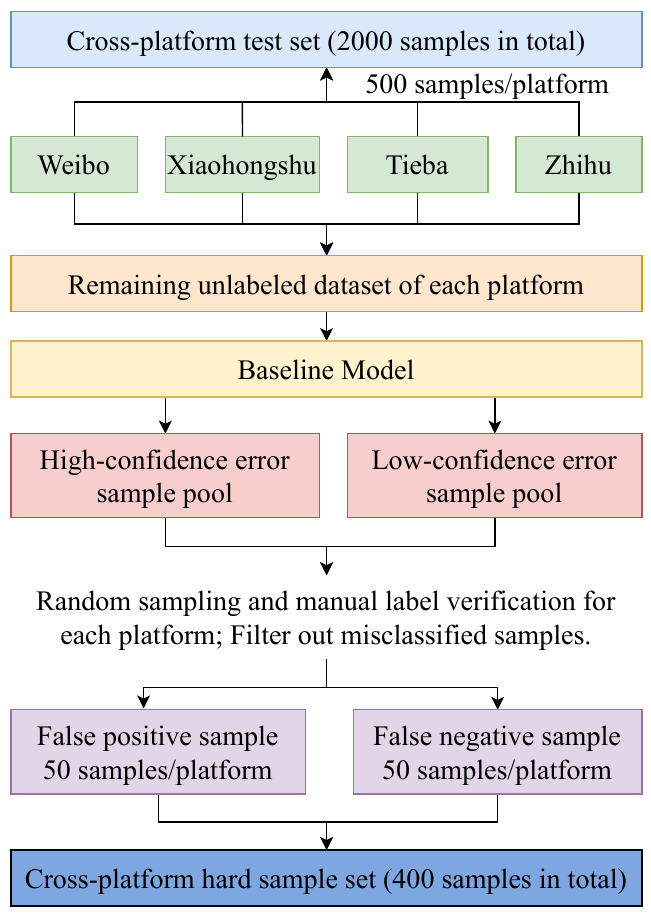}}
	\caption{Workflow of cross-platform hard sample mining with dual thresholds}
	\label{fig6}
\end{figure}

The overall screening pipeline is illustrated in Fig. \ref{fig6}. For each platform, 500 texts are reserved to form an independent test set, while the remaining candidate corpus is fed into the baseline detector for automatic binary classification. After that, high-confidence and low-confidence error pools are constructed separately following the dual-threshold rules. After stratified sampling, manual re-screening and verification are carried out. We fix 100 hard examples for each platform, including 50 false positive samples and 50 false negative samples with a 1:1 ratio of positive and negative hard cases. A total of 400 cross-platform hard samples aggregated from four platforms are adopted for secondary fine-tuning.

\textbf{(2) Implicit Context Model Calibration}

Secondary supervised fine-tuning is conducted on the 400 screened cross-platform hard samples to specifically compensate recognition blind spots of the baseline model on multi-platform implicit offensive content and niche community slang. The 400 hard samples are stratified and randomly split at a ratio of 8:2 to form the fine-tuning training set and independent validation set. A balanced 1:1 ratio of positive and negative samples is strictly maintained during stratification. Incremental fine-tuning is implemented on the aforementioned baseline detector. Hyperparameters are adapted according to the characteristics of few-shot fine-tuning, with detailed configurations listed in Table \ref{tab9}.

\begin{table}[t]
	\centering
	\caption{Hyperparameter Configuration for Fine-tuning}
		\renewcommand{\arraystretch}{1.3}
	\begin{tabular}{lc}
		\hline
		Hyperparameter & Value \\
		\hline
		Epoch & 10 \\
		Per-device train batch size & 16 \\
		Base learning rate & $2\times 10^{-5}$ \\
		Weight decay & 0.01 \\
		Evaluation strategy &  per epoch \\
		$\alpha$, $\beta$  &  0.85, 0.60\\
		\hline
	\end{tabular}
	\label{tab9}
\end{table}

Classification cross-entropy loss is calculated on the validation set upon the completion of each training epoch. A cross-platform optimized model is finally generated through the implicit context fusion calibration pipeline for subsequent horizontal comparison of generalization performance.

\subsection{Comparative Analysis of Optimized and Baseline Model}

Offline inference is performed on four independent test sets excluded from fine-tuning for the optimized model calibrated via hard sample fine-tuning. Horizontal performance comparison with the baseline model is conducted.

\begin{table}[t]
	\centering
	\renewcommand{\arraystretch}{1.3}
	\caption{Statistics of Cross-platform Generalization Performance Before and After Context Calibration}
	\begin{tabular}{llcccc}
		\hline
		Platform & Model & Accuracy & Precision & Recall & F1-Score \\
		\hline
		Weibo & Baseline & 0.708 & 0.591 & 0.337 & 0.430 \\
		& Optimized & 0.652 & 0.474 & 0.613 & 0.535 \\
		Xiaohongshu & Baseline & 0.652 & 0.444 & 0.143 & 0.216 \\
		& Optimized & 0.630 & 0.454 & 0.494 & 0.473 \\
		Tieba & Baseling & 0.592 & 0.613 & 0.490 & 0.545 \\
		& Optimized & 0.654 & 0.632 & 0.731 & 0.678 \\
		Zhihu & Baseline & 0.618 & 0.595 & 0.530 & 0.561 \\
		& Optimized & 0.678 & 0.641 & 0.683 & 0.661 \\
		\hline
	\end{tabular}
	\label{tab10}
\end{table}

As shown in Table \ref{tab10}, obvious improvements on F1 score and recall of harmful samples across four platforms are observed after implicit context calibration. The performance gain on Xiaohongshu dataset is the most remarkable with nearly doubled metrics, which demonstrates that targeted hard sample calibration brings prominent effectiveness for brand-new platforms not covered by COLD dataset and with severe domain shift. In terms of recall, substantial growth in the recall of offensive samples is observed across all four platforms, which indicates that the missing alarm issue of the baseline model against implicit offensive content is significantly alleviated by the optimized model.

\begin{figure}[t]
	\centerline{\includegraphics[width=3.3in,keepaspectratio]{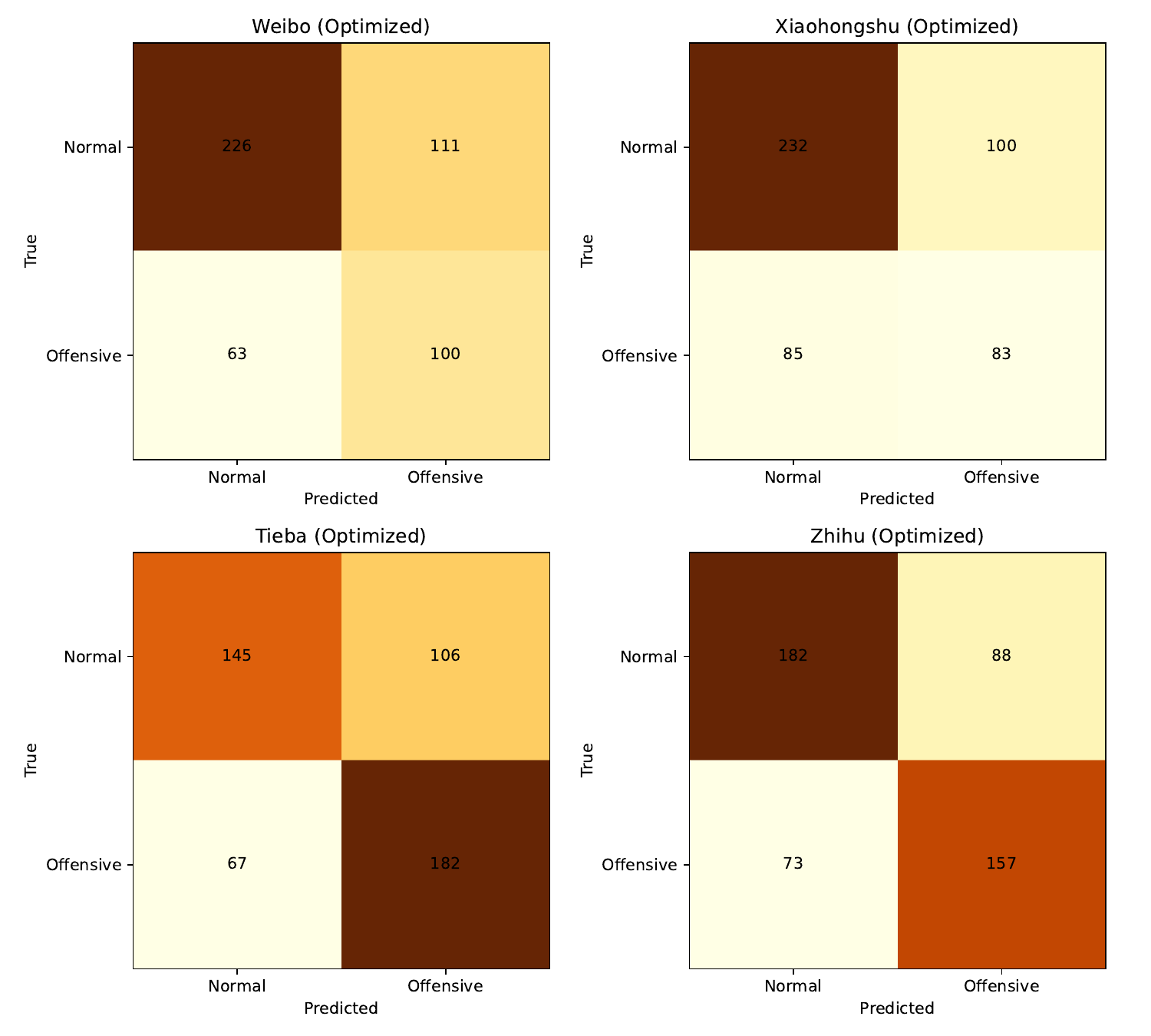}}
	\caption{Confusion Matrices for Classification Diagnosis of the Optimized Model on Each Platform}
	\label{fig7}
\end{figure}

The above performance variation can be further verified from the confusion matrix in Fig. \ref{fig7}. Taking the Xiaohongshu dataset as an example, the number of missed implicit offensive samples of the baseline model drops from 144 to 85, and the false negative rate decreases from 85.7\% to 50.6\%. Reductions in missed samples are also observed on other platforms. Substantial improvement in the recognition capacity for implicit offensive content such as irony and metaphor is validated for the optimized model. 

It is worth noting that a slight decline in precision is observed on partial platforms while the recall rate of harmful samples rises substantially. This phenomenon corresponds to a reasonable performance trade-off introduced when the classification decision boundary is moderately relaxed to reduce false negatives of offensive comments.

\begin{table}[htbp]
	\centering
		\renewcommand{\arraystretch}{1.3}
	\caption{Cross-platform Macro F1 under Different Sampling Strategies (Mean $\pm$ Std of 5 Runs)}
	\begin{tabular}{|c|c|c|c|}
		\hline
		Strategy & Budget & Macro F1 & Improvement vs. Baseline \\
		\hline
		Baseline & 0 & 0.438 & --- \\
		\hline
		Random-320 & 320 & $0.372\pm0.004$ & -0.066 \\
		\hline
		Hard-320 & 320 & $0.540\pm0.017$ & +0.102 \\
		\hline
	\end{tabular}
	\label{tab11}
\end{table}
\subsection{Ablation Study}

Three control groups are set to verify that performance gain originates from hard example mining rather than simple expansion of 320 data samples: (1) Baseline without fine-tuning; (2) Random-320 with 320 randomly sampled data for fine-tuning; (3) Hard-320 with 320 hard examples mined via dual thresholds for fine-tuning. Consistent annotation sources are guaranteed, as samples of Hard-320 and Random-320 are labeled by the same annotator following unified labeling standards.

Experimental results are presented in Table \ref{tab11}. The macro F1 of Hard-320 reaches $0.540\pm0.017$, which is significantly higher than $0.372\pm0.004$ of Random-320. It is demonstrated that hard example mining outperforms random sampling remarkably under identical annotation budget. It is notable that the metric of Random-320 is even lower than the untuned Baseline. This is attributed to the fact that multi-epoch fine-tuning on such limited samples leads to severe overfitting to the local distribution of target subsets, which impairs discriminative capabilities learned from the source domain. The positive gain brought by fine-tuning fails to offset the damage, resulting in degraded overall performance. In contrast, every hard sample in Hard-320 carries explicit correction signals, and remarkable performance gains are thereby achieved.

In addition, the proportions of real offensive content in the two original groups are 44.4\% and 29.1\%, respectively. After downsampling both groups to identical size and class ratio (93 positive and 178 negative samples for each) and rerunning experiments, the macro F1 of Hard group ($0.478\pm0.024$) remains higher than that of Random group ($0.413\pm0.011$). It is verified that the superiority stems from sample difficulty rather than class proportion.

\begin{figure}[t]
	\begin{minipage}{0.48\linewidth}
		\centerline{\includegraphics[width=\textwidth]{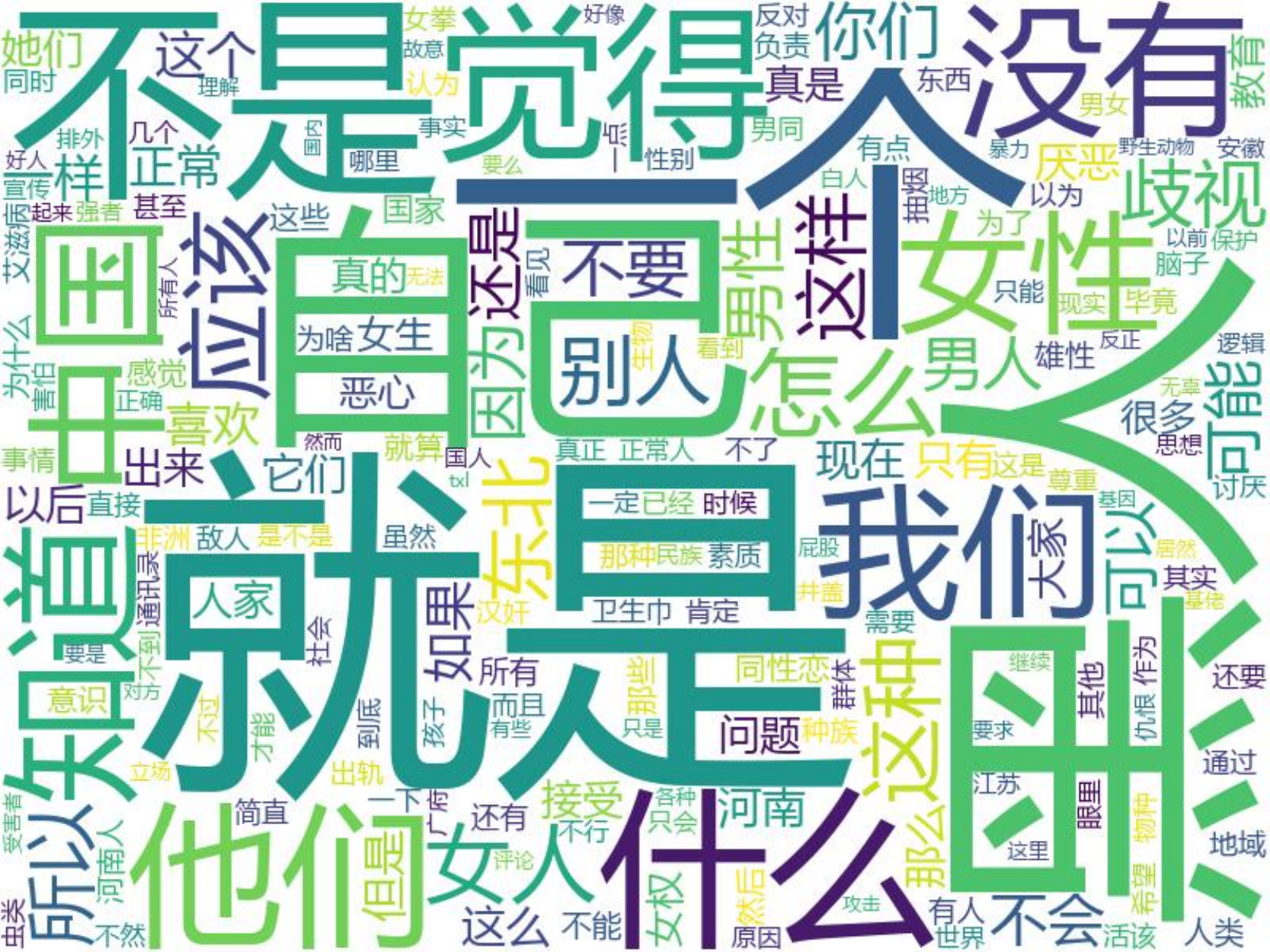}}
	\end{minipage}
	\vspace{3pt}
	\begin{minipage}{0.48\linewidth}
		\centerline{\includegraphics[width=\textwidth]{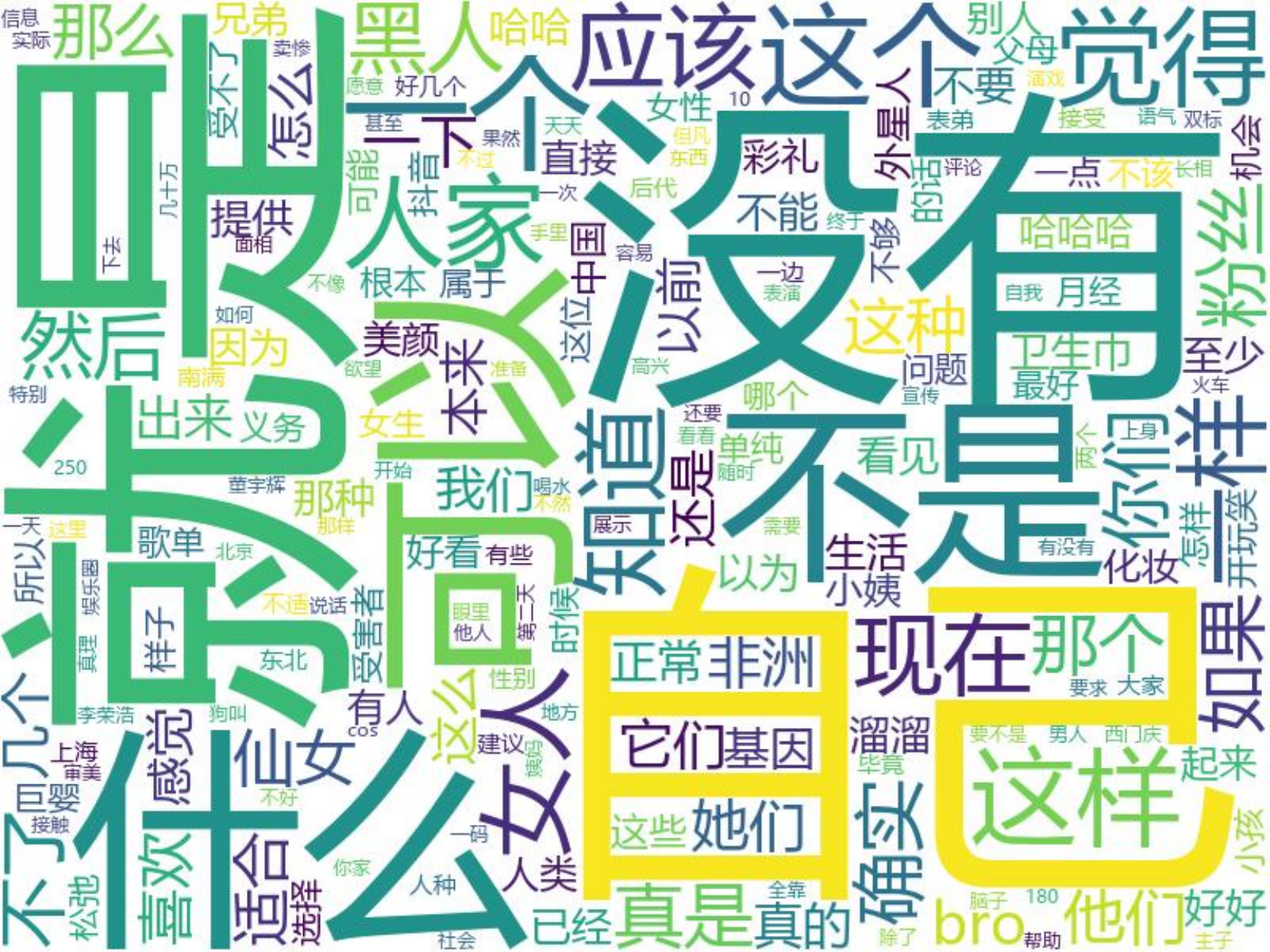}}
	\end{minipage}
	\caption{Comparison of Cross-platform Feature Word Clouds of Two Models (Left: Baseline; Right: Optimized)}
	\label{figs7}
\end{figure}

\subsection{Feature Word Clouds and Typical Cases}
To intuitively compare shifts in model attention before and after optimization from the feature perspective, high-frequency feature words of samples predicted as offensive by the two-stage models are extracted respectively to generate feature word clouds, as illustrated in Fig. \ref{figs7}.

Among the high-frequency features extracted from both models, besides explicit group and topic words such as discrimination and women, sentence structural words and oral function words including “\begin{CJK}{UTF8}{gbsn}就是\end{CJK}[just]”, “\begin{CJK}{UTF8}{gbsn}觉得\end{CJK}[think]”, “\begin{CJK}{UTF8}{gbsn}应该\end{CJK}should”, “\begin{CJK}{UTF8}{gbsn}什么\end{CJK}[what]” also occupy prominent weights. It indicates that model discrimination does not merely rely on matching a small set of sensitive words, but also pays considerable attention to sentence patterns and contextual structures where words are located. A comparison between the two word clouds reveals increased weights for literally neutral words frequently adopted for ironic or metaphorical attacks, such as "\begin{CJK}{UTF8}{gbsn}仙女\end{CJK}[fairy]", "\begin{CJK}{UTF8}{gbsn}外星人\end{CJK}[alien]" and "\begin{CJK}{UTF8}{gbsn}好看\end{CJK}[good-looking]", in the word cloud of the optimized model. This shift aligns well with the research target of calibrating the model with hard examples to detect implicit offensive content.

It should be clarified that the two word clouds are generated from distinct sample sets labeled as offensive by each model respectively. Thus, the visualization only serves as illustrative qualitative observation, and all valid conclusions are subject to the quantitative metrics presented above.

\begin{table}[t]
	\centering
	\renewcommand{\arraystretch}{1.3}
	\caption{Comparison of Two Models on Typical Ironic and Metaphorical Samples}
	\begin{tabular}{c p{4cm} cc}
		\hline
		Platform & Typical Implicit Offensiveness & Baseline & Optimized\\
		\hline
		Tieba & \begin{CJK}{UTF8}{gbsn}都怪我我应该送花圈。这样以后也能用得上\end{CJK} & 0 & 1\\
		Weibo & \begin{CJK}{UTF8}{gbsn}那你知道自己最后还是要嘎掉的, 怎么不天天随身携带棺材\end{CJK} & 0 & 1\\
		\hline
	\end{tabular}
	\label{tab12}
\end{table}

\begin{table}[t]
	\centering
	\caption{Comparison of Recognition of Context-Sensitive Expressions by Two Models}
	\renewcommand{\arraystretch}{1.3}
	\begin{tabular}{c p{3.8cm}cc}
		\hline
		Word & Text & Baseline & Optimized\\
		\hline
		\begin{CJK}{UTF8}{gbsn}普通\end{CJK} & \begin{CJK}{UTF8}{gbsn}这是普通民众朴素的愿望\end{CJK} & 0 & 0 \\
		\begin{CJK}{UTF8}{gbsn}普通\end{CJK} & \begin{CJK}{UTF8}{gbsn}...如果出现在国男身上，就不是优点了，毕竟这些特征那么普通又那么自信\end{CJK} & 0 & 1 \\
		\begin{CJK}{UTF8}{gbsn}好看\end{CJK} & \begin{CJK}{UTF8}{gbsn}ai 漫剧挺好看的，但 ai 真人是真难看\end{CJK} & 0 & 0 \\
		\begin{CJK}{UTF8}{gbsn}好看\end{CJK} & \begin{CJK}{UTF8}{gbsn}豆包说你好看，可能长在它的审美上\end{CJK} & 0 & 1 \\
		\begin{CJK}{UTF8}{gbsn}保护\end{CJK} & \begin{CJK}{UTF8}{gbsn}保护我方输出\end{CJK} & 0 & 0 \\
		\begin{CJK}{UTF8}{gbsn}保护\end{CJK} & \begin{CJK}{UTF8}{gbsn}不要出生是对她们最好的保护\end{CJK} & 0 & 1 \\
		\hline
	\end{tabular}
	\label{tab13}
\end{table}
\subsection{In-depth Analysis of Representative Recovered Cases}

To investigate improvements in semantic comprehension brought by implicit context calibration at the micro level, two types of typical samples are selected from the cross-platform test set for comparison of prediction results between the baseline and optimized model.

The first category consists of samples with ironic and metaphorical expressions, as shown in Table \ref{tab12}. Case 1 adopts funeral-related imagery such as “\begin{CJK}{UTF8}{gbsn}花圈\end{CJK}[wreath]” and “\begin{CJK}{UTF8}{gbsn}用得上\end{CJK}[will come in handy]” for implicit sarcasm, while Case 2 carries out personal attacks via coffin metaphors. Neither sample contains explicit abusive words on the surface. The baseline model relies on superficial keyword matching and misclassifies these samples as non-offensive, while the optimized model captures correlated semantics between imagery and sentence patterns to accurately identify offensive content.

The second category refers to context-sensitive expressions whose meaning heavily relies on context, as presented in Table \ref{tab13}. Three pairs of samples with identical literal words yet opposite sentiment polarities are selected, covering the terms “\begin{CJK}{UTF8}{gbsn}普通\end{CJK}[ordinary]”, “\begin{CJK}{UTF8}{gbsn}好看\end{CJK}[good-looking]” and “\begin{CJK}{UTF8}{gbsn}保护\end{CJK}[protect]”. Both models correctly label neutral usages as non-offensive. However, for implicit derogatory usages formed by contextual information, the baseline model suffers interference from the inherent neutral prior of these words and misses all such offensive samples. The optimized model dynamically distinguishes word semantics based on context and delivers correct predictions.

In summary, implicit context calibration shifts the model’s discrimination logic from reliance on local sensitive keywords to integrating sentence patterns and contextual semantic correlations, endowing it with the capability to detect implicit harmful speech such as irony, analogy and metaphor.

\section{Conclusion}\label{5}
This paper constructed a reproducible and comparable baseline model for offensive text detection based on chinese-roberta-wwm-ext. On this basis, a four-platform three-class cross-platform evaluation dataset was constructed. By adopting domain distance metrics including Proxy-A-distance, two primary causes underlying the cross-platform performance degradation of the baseline model were revealed. To tackle these bottlenecks, this paper proposed an optimization scheme combining dual-threshold hard example mining and secondary fine-tuning with implicit context. The approach achieved remarkable generalization performance across the four platforms with minimal labeling budget, and the performance gain was most prominent on Xiaohongshu, which suffered the largest domain gap. Ablation experiments demonstrated that performance gains stem from sample difficulty rather than sample quantity. After controlling the labeling budget and class distribution ratio, hard example mining consistently outperformed random sampling. Furthermore, this paper verified the reliability of the evaluation dataset via annotation consistency analysis and error case auditing.

\bibliographystyle{ieeetr} 
\bibliography{MyRefs} 
~~~\\
~~~\\

\end{document}